\documentclass{article} 

\usepackage{amsmath,amsfonts,bm}

\def\eqref#1{equation~\ref{#1}}

\def\1{\bm{1}}

\def\vg{{\bm{g}}}

\def\vm{{\bm{m}}}

\def\vv{{\bm{v}}}
\def\vw{{\bm{w}}}

\DeclareMathAlphabet{\mathsfit}{\encodingdefault}{\sfdefault}{m}{sl}
\SetMathAlphabet{\mathsfit}{bold}{\encodingdefault}{\sfdefault}{bx}{n}

\usepackage{microtype}
\usepackage{graphicx}
\usepackage{booktabs} 

\usepackage{hyperref}

\usepackage[accepted]{icml2020}

\usepackage{times}
\usepackage{url}
\usepackage{caption}
\usepackage{subcaption}
\usepackage{cleveref}
\usepackage{multirow}

\icmltitlerunning{NovoGrad - Stochastic Gradient Normalized by Layerwise Adaptive Second Moments}
\begin{document}

\twocolumn[
\icmltitle{Training Deep Networks with Stochastic Gradient Normalized by Layerwise Adaptive Second Moments}

\icmlsetsymbol{equal}{*}

\begin{icmlauthorlist}
\icmlauthor{Boris Ginsburg}{nv}
\icmlauthor{Patrice Castonguay}{nv}
\icmlauthor{Oleksii Hrinchuk}{nv}
\icmlauthor{Oleksii Kuchaiev}{nv}
\icmlauthor{Ryan Leary}{nv}
\icmlauthor{Vitaly Lavrukhin}{nv}
\icmlauthor{Jason Li}{nv}
\icmlauthor{Huyen Nguyen}{nv}
\icmlauthor{Yang Zhang}{nv}
\icmlauthor{Jonathan M. Cohen}{nv}
\end{icmlauthorlist}

\icmlaffiliation{nv}{NVIDIA, Santa Clara, USA}
\icmlcorrespondingauthor{\\ Boris Ginsburg}{bginsburg@nvidia.com}

\icmlkeywords{Deep Learning, Stochastic Gradient, NovoGrad}

\vskip 0.3in
]
\printAffiliationsAndNotice{\icmlEqualContribution} 

\begin{abstract}
We propose NovoGrad, an adaptive stochastic gradient descent method with layer-wise gradient normalization and decoupled weight decay. In our experiments on neural networks for image classification, speech recognition, machine translation, and language modeling, it performs on par or better than well-tuned SGD with momentum, Adam, and AdamW. Additionally, NovoGrad (1) is robust to the choice of learning rate and weight initialization, (2) works well in a large batch setting, and (3) has half the memory footprint of Adam.
\end{abstract}

\section{Introduction}
The most popular algorithms for training Neural Networks (NNs) are Stochastic Gradient Descent (SGD) with momentum \citep{Polyak1964, sutskever2013} and Adam \citep{Kingma2015}. SGD with momentum is the preferred algorithm for computer vision, while Adam is more commonly used for natural language processing (NLP) and speech problems. Compared to SGD, Adam is perceived as safer and more robust to weight initialization and learning rate.
However, Adam has certain drawbacks. First, as noted in the original paper~\citep{Kingma2015}, the second moment can vanish or explode, especially during the initial phase of training. 
Also, Adam often leads to solutions that generalize worse than SGD \citep{wilson2017}, e.g. models for image classification trained with Adam have significantly lower accuracy than when they are trained with SGD with momentum \citep{loshchilov2018}.

Our motivation for this work was to build an algorithm which: (1) performs equally well for image classification, speech recognition, machine translation, and language modeling, (2) is robust to learning rate (LR) and weight initialization, (3) has strong regularization properties. 

We started with Adam, and then (1) replaced the element-wise second moment with the layer-wise moment, (2) computed the first moment using gradients normalized by layer-wise second moment, (3) decoupled weight decay (WD) from normalized gradients (similar to AdamW). 

The resulting algorithm, \textit{NovoGrad}, combines SGD's and Adam's strengths. We applied NovoGrad to a variety of large scale problems --- image classification, neural machine translation, language modeling, and speech recognition --- and found that in all cases, it performs as well or better than Adam/AdamW and SGD with momentum.

\section{Related Work}
NovoGrad belongs to the family of \textbf{Stochastic Normalized Gradient Descent (SNGD}) optimizers~\citep{nesterov1984, hazan2015}. SNGD uses only the direction of the stochastic gradient $\vg_t$ to update the weights $\vw_t$:
\begin{equation*}
  \vw_{t+1} = \vw_t-\lambda_t \cdot\frac{\vg_t}{||\vg_t||}
\end{equation*}
The step size does not depend on the magnitude of that gradient. \citet{hazan2015} proved that the direction of the gradient is sufficient for convergence. Ignoring the gradient magnitude makes SNGD robust to vanishing and exploding gradients, but SNGD is still sensitive to "noisy" gradients, especially during an initial training phase. 

One can improve SNGD stability by gradient averaging. Adagrad \citep{duchi2011}, RmsProp \citep{Tieleman2012}, Adam \citep{Kingma2015} -- all these SNGD methods use the some kind of  gradient averaging. Adam, the most popular one, uses moving averages $\vm_t, \vv_t,$:
\begin{align}
 &\vm_t=\beta_1 \cdot \vm_{t-1} + (1-\beta_1)\cdot \vg_t\\
 &\vv_t=\beta_2 \cdot \vv_{t-1} + (1-\beta_2)\cdot \vg_t^2
\end{align}
The weights update is computed with the first moment $m_t$ normalized by the second moment $v_t$:
\begin{equation}
   \vw_{t+1} = \vw_t - \lambda_t \cdot \frac{\vm_t}{\sqrt{\vv_t}+\epsilon}
\end{equation}
where $\epsilon \ll 1$ is added for numerical stability. To strengthen Adam's robustness to "noisy" gradients, coefficients  $\beta_1$ and $\beta_2$ are usually close to $1$ (e.g. $\beta_2>0.99$). 

\subsection{Layer-wise Gradient Normalization}
The other way to improve SNGD robustness was proposed by \citet{yu2017}, who suggested to combine Adam with \textbf{layer-wise gradient normalization}. Both moments are computed with normalized gradients   $\hat{\vg}^l_t = \dfrac{\vg^l_t }{||\vg^l_t||}$, where $\vg^l_t$ is the gradient for the layer $l$ at step $t$:
\begin{align*} 
   &\vm^l_t = \beta_1 \cdot \vm^l_{t-1} + (1-\beta_1) \cdot \hat{\vg}^l_t \\
   &v^l_t = \beta_2 \cdot v^l_{t-1} + (1-\beta_2) \cdot ||\hat{\vg}^l_t||^2
\end{align*}
A similar idea was used in  \citep{singh2015} to scale up small layers gradients, while keeping large gradients unchanged:
\begin{align*} 
  \hat{\vg}^l_t = \vg^l_t \cdot (1 + \log(1 + \frac{1}{||\vg^l_t||}))
\end{align*}

\subsection{Improving Adam Generalization}
Adaptive methods like Adam generalize worse than SGD with momentum \citep{wilson2017}. For example,~\citet{keskar2018} proposed to use Adam during the initial stage only and then switch to SGD. \citet{luo2019} suggested to improve Adam generalization by limiting the factor $\frac{1}{\sqrt{v_t}}$ to a certain range: limiting from above helps to decrease the training loss while limiting from below helps to generalize better.

To improve Adam regularization, \citet{loshchilov2018}  proposed \textbf{AdamW}, which \textbf{decouples the weight decay} $d \cdot \vw_t$ from the gradient and uses it directly in the weight update:
\begin{equation*}
   \vw_{t+1} = \vw_t - \lambda_t \cdot (\dfrac{\vm_t}{\sqrt{v_t}+\epsilon} + d \cdot \vw_t)
   \label{eq:adamw}
\end{equation*}

\subsection{Reduction of Adam Memory Footprint}
Adam needs to store the second moment, and this doubles the optimizer memory compared to SGD with momentum. This affects large models like GPT-2 -- $1.5$ billion parameters ~\citep{radford2019language}, Meena -- 2.6 billion parameters \citep{Adiwardana2020}, or Megatron-LM -- 8.3 billion parameters \citep{Shoeybi2019}.
\citet{shazeer2018} proposed the \textbf{AdaFactor} algorithm, which replaced the full second moment with moving averages of the row and column sums of the squared gradients. For a layer defined by an $n\times m$ matrix, this would reduce memory from $\mathcal{O}(n \times m)$ to $\mathcal{O}(n + m)$.

\section{Algorithm}
\label{sec:algorithm}
NovoGrad combines three ideas: (1) use layer-wise second moments, (2) compute first moment with gradients normalized with layer-wise second moments, (3) decouple weight decay.

\begin{algorithm}[htb!]
\begin{algorithmic}
\STATE {\bf Parameters:} \\
\ \ Init learning rate $\lambda_0$,  moments $\beta_1, \beta_2$, weight decay $d$,\\
\ \ number of steps $T$
\STATE $t=0$: {\bf weight initialization}\\ \ \ $\vw_0 \gets Init()$.  
\STATE $t=1$: {\bf moment initialization}
 \FOR{ each layer $l$}
     \STATE $v^l_1   \gets ||\vg^l_1||^2;$
     \STATE $\vm^l_1 \gets \dfrac{\vg^l_1}{\sqrt{v^l_1}} + d \cdot \vw^l_1$.
 \ENDFOR
\WHILE {$ t \leq T$ }
\STATE compute global learning rate
   \ $\lambda_t\gets LR(\lambda_0,t,T)$ 
    \FOR{ each layer $l$}
        \STATE $\vg_t^l  \gets \nabla_l L(w_t)$  
        \STATE $v^l_t  \gets \beta_2 \cdot v^l_{t-1} + (1-\beta_2) \cdot ||\vg^l_t||^2$  
        \STATE  $\vm^l_t \gets \beta_1 \cdot \vm^l_{t-1} +  (\dfrac{\vg^l_t}{\sqrt{v^l_t} +\epsilon} + d \cdot \vw^l_{t})$
        \STATE $\vw^l_{t+1} \gets \vw^l_t - \lambda_t \cdot \vm^l_{t}$ 
    \ENDFOR
\ENDWHILE
\end{algorithmic}
\caption{NovoGrad \label{algo:NovoGrad}}
\end{algorithm}

Let $\vg^l_t$ be the stochastic gradient for layer $l$ at step $t$. First, we compute the layer-wise second moment $v^l_t$ using $||\vg^l_t||$:
\begin{equation}
    v^l_t = \beta_2  \cdot v^l_{t-1} + (1-\beta_2)  \cdot ||\vg^l_t||^2
\end{equation}
where $0 \leq \beta_2 \leq 1$. We  use much smaller $\beta_2$ than in Adam,\footnote{The default $\beta_2=0.25$ which we used in all experiments except Imagenet classification where $\beta_2=0.98$ was used.}
The moment $v^l_t$ is  used to normalize the gradient $\vg^l_t$ before calculating the first moment $\vm^l_t$. 
\begin{equation}
 \vm^l_t = \beta_1 \cdot \vm^l_{t-1} +  \dfrac{\vg^l_t}{\sqrt{v^l_t} +\epsilon} 
 \vw^l_{t}
\end{equation}
Last, we decouple weight decay $d\cdot \vw_t$ from the stochastic gradient similarly to AdamW, but we add it to normalized gradient before computing moment $\vm^l_t$:\footnote{One can also use decoupled weight decay in the weights update, as in AdamW. We didn't find a significant difference between these two options.}
\begin{equation}
    \vm^l_t = \beta_1 \cdot \vm^l_{t-1} +  (\dfrac{\vg^l_t}{\sqrt{v^l_t}+\epsilon} + d \cdot \vw^l_{t}
    \label{eq:novograd_w})
\end{equation}
where $0<\beta_1<1$ is the momentum, typically in the same range as in SGD or Adam $[0.9-0.95]$.
The first moment can be also computed with an exponential moving average in Adam-like style:
\begin{equation*}
\vm^l_t = \beta_1  \cdot \vm^l_{t-1} + (1-\beta_1) \cdot (\dfrac{\vg^l_t}{\sqrt{v^l_t}+\epsilon} + d \cdot \vw^l_{t})
\end{equation*}
We use the following moments initialization to remove bias:
\begin{align*}
   & v^l_1=||\vg^l_1||^2  \; ;  \; \;
     \vm^l_1= \dfrac{\vg^l_1}{||\vg^l_1||} + d \cdot \vw^l_1
\end{align*}
Finally, weights are updated the same way as in SGD with momentum:
\begin{equation*}
    \vw_{t+1} = \vw_t - \lambda_t \cdot \vm_t 
\end{equation*}
Similar to Adam, one can construct a counter-example for NovoGrad in the stochastic convex optimization settings \citep{wilson2017}. However, the ``AMS-Grad'' fix~\citep{reddi2018} for Adam can also be applied in this case to make sure that $\dfrac{\lambda_t}{\sqrt{v^l_t}}$ is monotonically decreasing:
\begin{align*} 
   &v^l_t = \beta_2  \cdot v^l_{t-1} + (1-\beta_2)  \cdot ||\vg^l_t||^2 \\
   &\hat{v}^l_t = \max (\hat{v}^l_{t-1}, v^l_t) \\
   &\vm^l_t = \beta_1  \cdot \vm_{t-1} + ( \dfrac{\vg^l_t}{\sqrt{\hat{v}^l_t}+\epsilon} + d \cdot \vw^l_{t})
  \label{eq:AMS-NovoGrad} 
\end{align*}
\textit{Notes}.
\begin{enumerate}
    \item If we set $\beta_2=0$, then $v^l_t = ||\vg^l_t||^2$, and NovoGrad becomes layer-wise NGD.
    \item We use  gradient normalization in the first moment computation instead of moment normalization in weights update to improve the algorithm robustness against very large "outliers" gradients.
    \item NovoGrad has half the memory footprint of Adam. 
\end{enumerate}

\subsection{Training with NovoGrad}
NovoGrad has initial learning rates different than both  SGD and Adam.  In the initial phase, normalized gradients have larger magnitudes than non-normalized gradients used by SGD, so NovoGrad uses smaller learning rate than SGD. In Adam, on the other side, normalization is done by element-wise second moments, which are significantly smaller than per-layer second moments used in NovoGrad. So for NovoGrad, safe learning rates are somewhere between those of SGD and Adam, as the gradients are normalized by the per-layer gradient norm.   

\section{Experiments}
We train four deep models:
ResNet-50~\citep{he2016} --- for ImageNet classification,
Transformer-big~\citep{vaswani2017} --- for WMT 2014 translation,
Jasper~\citep{li2019} --- for LibriSpeech speech recognition, and 
Transformer-XL~\citep{dai2019transformer} --- for WikiText-103 word-level language modeling, with NovoGrad, SGD with momentum, and Adam/AdamW.
Each model was trained on a single DGX-1 with $8$ NVIDIA V$100$ GPUs with gradient accumulation used for large batch training. In all the experiments, NovoGrad performed on par or better than other algorithms. 

\subsection{Image Classification}
We used ResNet-50v2 model~\cite{he2016} for ImageNet classification task~\cite{ILSVRC15}.
We trained the model with three optimizers: SGD with momentum (SGD), AdamW, and NovoGrad using the batch of 1024 for 100 epochs. 
We used quadratic LR decay for SGD with momentum and cosine decay~\citep{loshchilov2016} for AdamW and NovoGrad.
We could not find any training recipe for ResNet-50 with AdamW, so we report the best accuracy we achieved after extensive hyper-parameter search. We used only standard data augmentation methods: resize, flip, random crop, and did not employ any additional training tricks~\citep{he2018}. The single-crop validation accuracy for each algorithm is reported in \Cref{tab:ImageNet2}.

\begin{table}[!ht]
\centering
\caption{ImageNet: ResNet-50v2 trained with batch 1024, top-1/top-5 accuracy (\%).}
\label{tab:ImageNet2}
\begin{tabular}{ccccccc} 
 \toprule
  \textbf{Optimizer} &\textbf{LR}&\textbf{WD} &\textbf{Epoch}& 
  \textbf{Top-1/Top-5,\%}\\
 \midrule
  \multirow{2}{*}{SGD} & \multirow{2}{*}{$0.400$}  & \multirow{2}{*}{$0.0001$} & $100$ & $76.38/93.08$\\
      & &  & $200$ & $76.33/92.96$\\
 \midrule
  \multirow{2}{*}{AdamW}  & \multirow{2}{*}{$0.002$} & \multirow{2}{*}{$0.120$} & $100$ & $76.36/93.01$\\
        & & &  $200$ & $76.48/92.94$\\
  \midrule
  \multirow{3}{*}{NovoGrad} &  \multirow{3}{*}{$0.007$} & \multirow{3}{*}{$0.002$} & $100$ & $76.94/93.41$\\
  & &  & $200$ & $77.74/93.70$\\ 
  & &  & $300$ & $77.65/93.62$\\  
  \bottomrule
\end{tabular}
\end{table}
\begin{table*}[t]
\centering
\caption{ImageNet, large batch training comparison: ResNet-50v2, top-1 accuracy(\%).}
\label{tab:LargeBatch}
\begin{tabular}{cccccc} 
 \toprule
\textbf{Optimizer} & \textbf{Reference} & \textbf{Bag of Tricks} & \textbf{Epochs} & \textbf{B=8K} & \textbf{B=32K} \\
 \midrule
  SGD & \cite{goyal2017} & warmup & 90 & 76.26 & 72.45 \\
  SGD & \cite{you2018} & warmup, LARS  & 90 & 75.30 & 75.40  \\ 
  SGD & \cite{Codreanu2017} & warmup, multi-step WD & 100 & 76.60 & 75.31  \\
  \midrule 
   \multirow{2}{*}{NovoGrad} & & ---  & 90 & 76.64 & 75.78 \\
   &           & warmup  & 90 & --- & 75.99 \\
  \bottomrule
\end{tabular}
\end{table*}
NovoGrad outperformed both AdamW and SGD with the top-1 accuracy of 76.94\% after 100 epochs. SGD and Adam accuracy remained under 76.5\% if we trained for 200 epochs instead, while NovoGrad accuracy improved to 77.74\%. NovoGrad demonstrated powerful regularization capabilities: training for 100 additional epochs kept top-1=77.65\% without overfitting.
Note that this is "vanilla" ResNet-50, without sophisticated data augmentation or model tweaking.

\subsection{Large Batch Training for Image Classification}
\citet{hazan2015} showed that large batch size is beneficial for SNGD convergence, which motivated us to explore NovoGrad for large batch training. We trained ResNet-50 v2 with batch sizes of 8K and 32K. To compare with the previous methods, we train the model for 90 epochs using cosine LR decay. To emulate a large batch, we used a mini-batch of 128 per GPU and accumulated gradients from several mini-batches before each weight update.

To establish the baseline for NovoGrad training with batch 32K we first used the method similar to proposed in~\citet{goyal2017}: scaling the learning rate linearly with the batch size and using LR warmup. This method gives top-1=$75.09\%$ and top-5=$92.27\%$. We found that we get better results when we increase both the learning rate $\lambda$ and the weight decay $d$ to improve the regularization (see \Cref{tab:LargeBatch_NovoGrad}).
\begin{table}[!ht]
\centering
\caption{ImageNet, large batch training with NovoGrad: ResNet-50v2, 90 epochs. top-1/top-5 accuracy (\%).}
\label{tab:LargeBatch_NovoGrad}
\begin{tabular}{ccccc} 
 \toprule
  \textbf{Batch} & \textbf{LR} & \textbf{WD} & \textbf{Top-1/Top-5,\%} \\
 \midrule
  1K & $0.070$  & $0.002$ & $76.86/93.31$ \\
  8K & $0.016$  & $0.006$ &  $76.64/93.14$ \\
  32K & $0.026$  & $0.010$ &  $75.78/92.54$ \\
  \bottomrule
\end{tabular}
\end{table}

For comparison, we took three methods, which (1) use fixed batch size during training and (2) do not modify the original model. All three methods employ SGD with momentum. The first method (\citet{goyal2017}) scales LR linearly with batch size and uses the LR warmup to stabilize the initial training phase. The second method (\citet{you2018}) combines warmup with Layer-wise Adaptive Rate Scaling (LARS) \citep{lars2017}. The last method (\citet{Codreanu2017}) uses warmup and dynamic weight decay (WD).
NovoGrad outperformed other methods without any additional techniques like LR warmup \citep{goyal2017}, dynamic weight decay, special batch norm, etc. Using warm-up (500 steps) improves top-1 accuracy to $75.99\%$.

\subsection{Speech Recognition}
We conducted experiments with Jasper DR 10x5  (\citet{li2019}), a deep convolutional neural acoustic model, on the LibriSpeech speech recognition task \citep{panayotov2015}. Jasper was trained with SGD with momentum (SGD), Adam and NovoGrad for $400$ epochs with a batch of $256$, polynomial LR decay, and  Layerwise Adaptive Rate Clipping (LARC).
We found that NovoGrad yields lower Word Error Rates (WER) comparing to SGD, especially for the long runs. The model and training parameters are described in \citet{li2019}.
\begin{table}[!th]
\centering
\caption{Speech Recognition: Jasper-10x5 trained on LibriSpeech for 400 epochs, greedy WER(\%).}
\label{tab:LibriSpeechResults}
\begin{tabular}{ccccc} 
 \toprule
  \multirow{2}{*}{\textbf{Optimizer}} & \multicolumn{2}{c}{\textbf{Dev}} & \multicolumn{2}{c}{\textbf{Test}} \\
  & {\textbf{clean}} & {\textbf{other}} & {\textbf{clean}} & {\textbf{other}} \\
\midrule
 Adam & $5.06$ & $15.76$ & $5.27$ & $15.94$ \\
 SGD  & $4.08$ & $13.23$ & $4.22$ & $13.15$ \\ 
\midrule
 NovoGrad & $3.71$ & $11.75$ & $3.71$ & $11.85$ \\ 
\bottomrule
\end{tabular}
\end{table}

\subsection{Large Batch Training for Speech Recognition}
\begin{table*}[!ht]
\centering
\caption{Speech Recognition, large batch training. Jasper-10x5 trained on LibriSpeech for 400 epochs, WER(\%).}
\label{tab:Jasper_LargeBatch_NovoGrad}
\begin{center}
\begin{tabular}{cccccccc} 
 \toprule
 \multirow{2}{*}{\textbf{Batch}} & \multirow{2}{*}{\textbf{LR}} & \multirow{2}{*}{\textbf{Warmup}} & \multirow{2}{*}{\textbf{WD}} & \multicolumn{2}{c}{\textbf{Dev}} & \multicolumn{2}{c}{\textbf{Test}} \\
  & & & & \textbf{clean} & \textbf{other} & \textbf{clean} & \textbf{other} \\
 \midrule
  $512$& $0.015$ & -   & $0.001$ &  $3.58$ & $11.30$ & $3.85$ & $11.29$ \\
  $4$K & $0.03$ & $0.05$ & $0.001$ &  $3.66$ & $11.67$ & $3.92$ & $11.68$ \\
  $8$K & $0.06$ & $0.05$ & $0.001$  &  $3.69$ & $11.76$ & $3.96$ & $11.75$ \\
  $16$K & $0.06$ & $0.05$ & $0.003$ &  $3.67$ & $11.03$ & $3.94$ & $11.19$ \\
  $32$K & $0.06$ & $0.08$ & $0.004$ &  $4.01$ & $11.73$ & $4.14$ & $11.89$ \\
  \bottomrule
\end{tabular}
\end{center}
\end{table*}
\begin{figure}[htb!]
  \centering
  \includegraphics[width=\columnwidth]{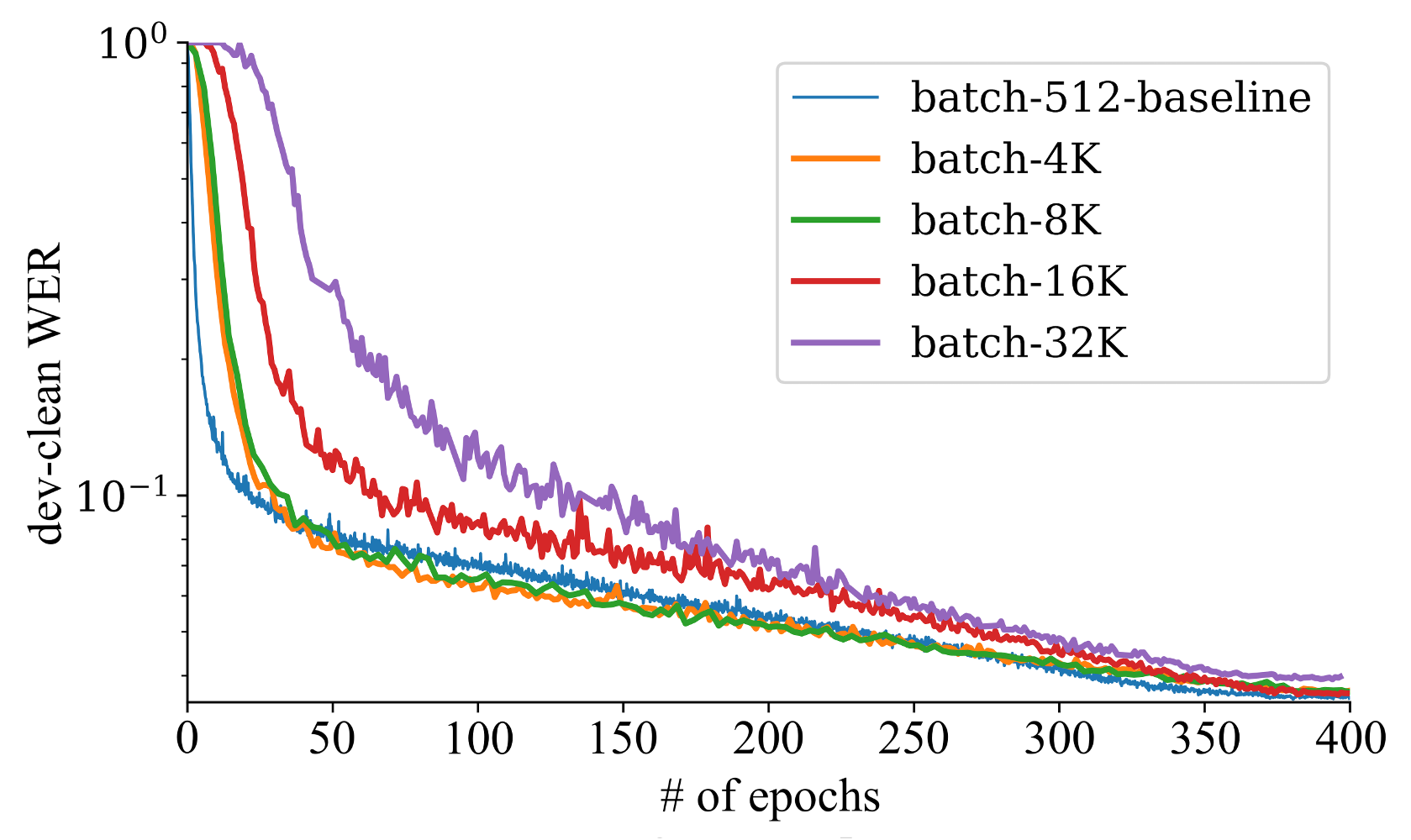}
  \captionof{figure}{Speech Recognition, large batch training. Jasper-10x5 trained with NovoGrad on LibriSpeech, WER(\%).}
  \label{fig:asr_learning_curve}
\end{figure}
We trained Jasper DR 10x5 with batch sizes of 512, 4K, 8K, 16K and 32K on LibriSpeech. In all cases, we trained the model for 400 epochs. For batch size up to 8K, we scaled LR linearly with the batch size and used LR warmup. To scale batch to 16K and 32K we also increased weight decay (see \Cref{tab:Jasper_LargeBatch_NovoGrad}). The batch 16K leads to WER comparable to the baseline. Batch 32K has higher WER due to the smaller number of training steps (9 weights updates per epoch). 
\Cref{fig:asr_learning_curve} shows WER on dev-clean during training for different batch sizes. 

\subsection{Neural Machine Translation}
We trained Transformer~\citep{vaswani2017} on WMT 2014 English-to-German benchmark. For all the experiments, we used a $12$-layer Transformer-big model with $185$M parameters ($d_\text{model}=1024$, $d_\text{ff}=4096$, $h=16$) with the vocabulary of $8192$ tokens based on joint source-target byte-pair-encodings~\citep{sennrich2015neural}. For Adam and AdamW we used dropout of $P_\text{drop}=0.3$ and for NovoGrad we used $P_\text{drop}=0.2$. We trained all algorithms with mixed-precision~\citep{micikevicius2017} for $100$K steps (approximately $150$ epochs) with a $4$K steps warmup on batches of up to $490$K source and target tokens obtained via gradient accummulation~\citep{ott2018scaling} with cosine learning rate annealing. We did not use checkpoint averaging, all the results are reported for the last checkpoint in the corresponding run.
\begin{table}[!ht]
\centering
\caption{WMT'14 English-to-German translation, Transformer-big,  batch $490$K tokens, $150$ epochs, no checkpoint averaging. Detokenized SacreBLEU and Tokenized BLEU on WMT'14 (newstest14).}
\label{tab:wmt_ende}
\begin{center} 
\begin{tabular}{ c c c c c c } 
\toprule
  \textbf{Optimizer}& \textbf{LR} & \textbf{WD} & \textbf{Sacre/Token-BLEU} \\
 \midrule 
  Adam  & $0.0006$  & - & $28.26/28.71$ \\
  AdamW & $0.0006$  & $0.005$ & $28.24/28.72$ \\
 \midrule
  NovoGrad & $0.04$ & $0.0001$ & $28.80/29.35$\\
\bottomrule
\end{tabular}
\end{center}
\end{table}

\subsection{Language Modeling}
\begin{table}[htb!]
\centering
\caption{Language Modeling. Transformer-XL trained on WikiText-103 with batch size $256$, sequence length $512$, 12B tokens. Validation and Test Perplexity(PPL).}
\label{tab:wikitext103}
\begin{center}
\begin{tabular}{cccccc}
\toprule
 {\textbf{Optimizer}} & {\textbf{LR}} & {\textbf{WD}} & {\textbf{Val /Test-PPL}} \\
\midrule
 Adam &  $0.00025$ & - & $23.84/25.40$  \\
 AdamW &  $0.00025$ & $0.001$& $23.64/25.06$ \\
\midrule
 NovoGrad &  $0.01$ & $0$ & $20.53/21.26$ \\
\bottomrule
\end{tabular}
\end{center}
\end{table}
We trained Transformer-XL~\citep{dai2019transformer}, the state-of-the-art language model on the word-level WikiText--103~\citep{merity2016pointer} benchmark. For all the experiments we used a $16$-layer model with $191$M parameters ($d_\text{model}=512$, $d_\text{ff}=2048$, $h=8$, $P_\text{drop}=0.15$). All other hyper-parameters were taken from the original Transformer-XL paper, the source code was based on a publicly available implementation.
Each configuration was trained for $12$ billion tokens which is approximately $117$ epochs and $366$K iterations.

\begin{figure}[htb!]
  \centering
  \includegraphics[width=\columnwidth]{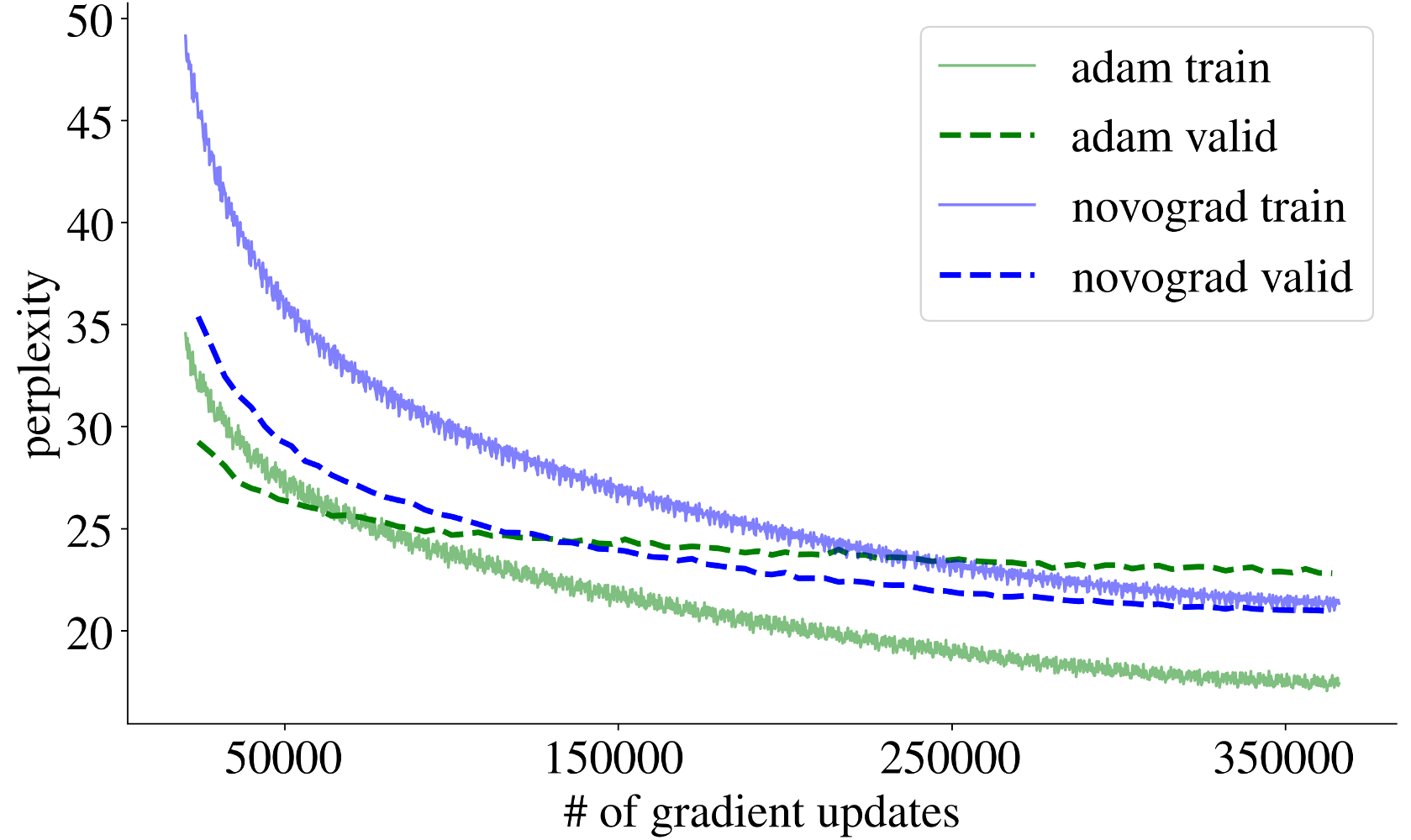}
  \captionof{figure}{Language Modeling. Transformer-XL trained with Adam and NovoGrad on WikiText-103.}
  \label{fig:txl_wt103}
\end{figure}

NovoGrad significantly outperformed both Adam and AdamW (\Cref{tab:wikitext103}). NovoGrad exhibits a much smaller gap between training and validation perplexity compared to Adam (\Cref{fig:txl_wt103}). Even longer training for $20$B tokens does not lead to overfitting, as the validation and test perplexities improve even further.

\section{Conclusion}
We propose NovoGrad -- an adaptive SGD method with gradients normalized by the layer-wise second moment and with decoupled weight decay. We tested NovoGrad on deep models for image classification, speech recognition, translation, and language modeling.

In all  experiments, NovoGrad performed equally or better than SGD and Adam/AdamW. NovoGrad is more robust to the initial learning rate and weight initialization than other methods. For example, NovoGrad works well without LR warm-up, while other methods require it. NovoGrad performs exceptionally well for large batch training, e.g. it outperforms other methods for ResNet-50 for all batches  up to 32K. In addition, NovoGrad requires half the memory compared to Adam. 

NovoGrad and all models in this paper are open sourced.

\bibliography{bibliography}
\bibliographystyle{icml2020}

\end{document}